\documentclass[letterpaper]{article} 
\usepackage{aaai25}  
\usepackage{times}  
\usepackage{helvet}  
\usepackage{courier}  
\usepackage[hyphens]{url}  
\usepackage{graphicx} 
\usepackage{natbib}  
\usepackage{caption} 
\usepackage{algorithm}
\usepackage{algorithmic}
\usepackage{multirow}
\usepackage{color}
\usepackage{newfloat}
\usepackage{listings}
\DeclareCaptionStyle{ruled}{labelfont=normalfont,labelsep=colon,strut=off} 
\floatstyle{ruled}
\newfloat{listing}{tb}{lst}{}
\floatname{listing}{Listing}
\title{Does Prompt Design Impact Quality of Data Imputation by LLMs?}

\author{Shreenidhi Srinivasan, 
Lydia Manikonda\textsuperscript{\rm 1}}
\affiliations{%
\textsuperscript{\rm 1}Corresponding author\\
Rensselaer Polytechnic Institute \\ Troy, New York, USA\\
\{srinis10, manikl\}@rpi.edu}

\usepackage{bibentry}

\begin{document}

\maketitle

\begin{abstract}
Generating realistic synthetic tabular data presents a critical challenge in machine learning. It adds another layer of complexity when this data contain class imbalance problems. This paper presents a novel token-aware data imputation method that leverages the in-context learning capabilities of large language models. This is achieved through the combination of a structured group-wise CSV-style prompting technique and the elimination of irrelevant contextual information in the input prompt. We test this approach with two class-imbalanced binary classification datasets and evaluate the effectiveness of imputation using classification-based evaluation metrics. The experimental results demonstrate that our approach significantly reduces the input prompt size while maintaining or improving imputation quality compared to our baseline prompt, especially for datasets that are of relatively smaller in size. The contributions of this presented work is two-fold -- 1) it sheds light on the importance of prompt design when leveraging LLMs for synthetic data generation and 2) it addresses a critical gap in LLM-based data imputation for class-imbalanced datasets with missing data by providing a practical solution within computational constraints. We hope that our work will foster further research and discussions about leveraging the incredible potential of LLMs and prompt engineering techniques for synthetic data generation. 
\end{abstract}

%

\section{Introduction}
Tabular data is one of the most common forms of data in machine learning. Over $65\%$ of datasets in the Google Dataset Search platform~\cite{GoogleDataset} contain tabular files in either CSV or XLS formats~\cite{benjelloun2020google}. Some of the major issues with tabular data are that they -- 1) are often class-imbalanced, 2) unavailable because of privacy concerns and 3) contain noisy or missing data. Performance of machine learning models depends on the quality and quantity of data they are trained on. Thus, research on techniques to alleviate the three important problems mentioned above have received considerable attention in the recent years. In this paper, \emph{we aim to develop a data imputation technique for class-imbalanced tabular datasets using Large Language Models (LLMs)}.

In several real-world classification problems, it is common for certain classes to have a significantly larger number of samples (majority classes), while others are underrepresented in the dataset (minority classes). This imbalance may stem from a lack of sufficient samples for the minority class or the high cost associated with obtaining such data. Consequently, a model trained on this type of dataset when deployed in the wild becomes biased and tends to perform poorly for the minority class. In addition to the above mentioned concerns about real-world tabular datasets, they can be complex to handle, in general. Tabular datasets typically contain both categorical and numerical features and hence, they require extensive preprocessing. Data preprocessing typically involves steps like encoding categorical data into numbers, data scaling or normalization and removing outliers. However, these actions could lead to the loss of crucial information or introduction of artifacts that weren't present in the original data.

Various strategies to deal with class-imbalanced datasets have been explored in the literature. Traditional solutions include undersampling the majority class and/or oversampling the minority class to balance class distribution, as well as hybrid sampling strategies that focus on selecting and retaining challenging samples while discarding those that are easier to learn. Additional methods include cost-sensitive analysis and synthetic data generation. The process of synthetic data generation for the minority class involves creating new samples corresponding to the minority class to augment the original dataset. Several techniques have been proposed for this purpose, including sampling methods like SMOTE~\cite{chawla2002smote}. With the rise of deep learning models, Variational Autoencoders~\cite{kingma2013auto} and Generative Adversarial Networks~\cite{goodfellow2014generative} have proven effective in generating realistic tabular data ~\cite{kim2024}. Notably, GReaT~\cite{borisov2022language} has surpassed previous methods in producing tabular data formatted in natural language. However, GReaT requires the fine-tuning of parameters in large language models, which can be resource-intensive, as extensive training is needed for each dataset ~\cite{kim2024}. Consequently, applying this technique across a variety of datasets and domains poses challenges.

Lately, there has been a notable advancement in prompt engineering research~\cite{brown2020,guo2023,kojima2022,wei2022,zhou2022} to leverage the capabilities of LLMs for specific tasks while mitigating additional training costs~\cite{kim2024}. Kim et al.~\cite{kim2024} proposes a novel group-wise prompting method presented in a CSV-style format to generate high-quality synthetic data for class-imbalanced datasets. This method exploits the in-context learning capabilities of LLMs to generate data that closely aligns with the desired characteristics of the target dataset. The method consistently enhances machine learning performance across eight public datasets while maintaining the integrity of data distributions and feature correlations. In this paper, we propose an analogous prompting strategy tailored for data imputation rather than synthetic data generation.

Our proposed method leverages the in-context learning abilities of LLMs using a structured CSV-style prompting technique designed to fill in the missing values in tabular datasets effectively. \emph{The core contribution of this paper is developing a procedure to maintain the quality of imputation while minimizing the input and output token size}. The imputation process focuses on one feature at a time utilizing information from its strongly correlated predictors, and this significantly reduces the input prompt length to accommodate more data samples within fixed token constraints. The prompt design omits weakly correlated or unrelated features that prevent the LLM from performing its assigned task effectively. Several research studies show that irrelevent context can significantly degrade the performance of LLMs ~\cite{jin2025end,jiang2024enhancing,shi2023large}. We measure the quality of imputation based on the classification performance of machine learning models built for two separate datasets obtained from Kaggle. We report the class-wise precision, recall, F1 and balanced accuracy scores. Our experimental results suggest that \emph{we successfully reduce the size of input prompt while maintaining the quality of imputation}. 

\noindent
\textbf{Related Work} Synthetic data is defined as an artificially generated dataset mimicking real world data where, in the context of machine learning, they are typically generated using algorithms or neural networks~\cite{nadas2025synthetic,park2018data}. Synthetic datasets are typically used in several domains and applications where, obtaining large amounts of training data is difficult due to several reasons. As machine learning models became more complex, they needed large quantities of high-quality training data that is also realistic. This led to using generative models to generate synthetic data~\cite{eigenschink2023deep,jiang2024generative,yu2023large,yu2023regen}. Also, several factors such as costs of collecting data via surveys or experiments, scarcity of data in certain domains, flexibility of scaling and controlling the underlying data distributions, preserving privacy, etc., led to a growing reliance on large language models to generate synthetic datasets~\cite{liu2024best,li2024data,long2024llms}. Building upon recent advancements in prompt engineering research for synthetic data generation, our work presented in this paper specifically focuses on data imputation using LLMs and investigates how prompt design will have an influence on the quality of data imputation for class-imbalanced datasets (especially for the minority class) as evaluated by classification metrics.

\section{Data and Preprocessing}
We test our approach with two publicly available tabular classification datasets, namely, the Adult Income dataset and the Travel dataset from Kaggle ~\cite{adult_income_kaggle,tour_churn_kaggle}. The feature descriptions for the prompt are directly sourced from Kaggle's data descriptions when available or we use ChatGPT to generate them. As shown in Table \ref{tab:Dataset}, the Adult Income dataset contains $48,842$ records with $14$ features. The target column for this dataset is `income' that takes two categorical values, $<=50$ K (majority class) and $>50$ K (minority class). The dataset contains $3620$ rows with missing values in the features `workclass', `occupation' and `native-country'. For our experiment, we consider $958$ rows with missing entries with $479$ records corresponding to each income class. The Travel dataset contains $954$ records with $6$ features. Its target column is `Target' which takes two values $0$ (customer does not churn)  and $1$ (customer churns). In this dataset, $1$ corresponds to the minority class. The original dataset contains a total of $60$ missing values in the `FrequentFlyer' feature column. For our experiment, we consider a synthetic version of this dataset obtained by introducing artificial missingness into the `FrequentFlyer' column to give a total of $120$ missing values ($60$ in each class). The missingness is introduced in randomly chosen records from each class.

Both these datasets are class-imbalanced with the minority class comprising about $24\%$ of the records. Apart from the domain differences of these two datasets, they significantly differ in terms of the total number of data points and total number of features. The Adult Income dataset is much larger ($48,842$ data points) compared to Travel dataset ($954$ records). Also, Adult Income has about double the number of features as compared to Travel as shown in Table~\ref{tab:Dataset}.

\begin{table}[h]
\begin{tabular}{|l|l|l|l|p{0.75in}|}
\hline Dataset&\#Class&\#Feats.&\#Samples&\#Incomplete samples\\
\hline Travel& 2 & 7 & 954 & 120\\
\hline Income& 2 & 15 & 48,842& 3,620\\
\hline
\end{tabular}
\caption{Summary of the datasets used in our work.}
\label{tab:Dataset}
\vspace{-0.55cm}
\end{table}
\section{Methodology}
In this study, we introduce an approach for tabular data imputation using LLMs. The missing entries are imputed using a CSV-style prompt carefully crafted to deal with class imbalanced datasets. Our objective is to fill in the missing values with a focus on minimizing input and output token usage. This study demonstrates the incredible potential of LLMs in the task of tabular data imputation. Inspired by an effective and token-efficient prompting technique used for synthetic data generation for imbalanced datasets~\cite{kim2024epic}, we adopt a similar group-wise prompting method to perform missing data imputation in imbalanced tabular datasets. In an effort to reduce the input token consumption further, we only include the features that hold a strong relationship with the feature that contains the missing entries to be imputed using the prompt. For datasets with multiple features with missing values, we impute one feature at a time. By including information from the most relevant feature columns only, we not only reduce the token size of the input prompt but also eliminate noisy data that can cause the LLM to learn patterns that do not accurately reflect the true dataset structure and relationships. We measure the effectiveness of imputation based on the classification performance of two ensemble learning models. 


\subsection{Group-wise Prompting Method}
The main idea is to construct prompts in a structured and predictable format, aiming to guide LLMs in synthesizing data entries maintaining the characteristics of the original dataset. This prompting method is designed to leverage the in-context learning capability of LLMs. Our goal is to surpass the bare enumeration of examples in the prompt, acknowledging the significance of this structured presentation in the production of high-quality realistic data entries. The prompt starts with an explicit instruction to the LLM to perform the imputation task by utilizing its in-context learning abilities. The instruction also tells the LLM to output the imputed feature column only (rather than the entire dataset) to reduce the output token size. This also helps in lowering the computational costs associated with the imputation process. In addition, the missing record sample size is explicitly stated to prevent the LLM from generating useless tokens after the response. The prompt continues with a set of brief one-line descriptions of features in the dataset to provide enhanced dataset context for imputation. The descriptions are followed by a header listing all the feature names in comma-separated format. The header is followed by predefined groups and each group consists of a fixed number of data samples. The groups are usually determined by the classes present in the target, for example, the groups for the Travel dataset would be \emph{Customer churns} and \emph{Customer does not churn}. The samples for each group are obtained by random sampling from complete records corresponding to that group in the dataset. This template is repeated with various samples from the dataset. Below the completed examples, the same number of records with missing values are presented in group-separated format and the LLM is instructed to impute the missing values based on its analysis of the complete samples.

\begin{table}[t]
\centering
\setlength{\tabcolsep}{3pt}
\small
\begin{tabular}{|l|ccc|l|}
\hline
\textbf{CSV Prompt Style} & \multicolumn{3}{c|}{\textbf{Income $\mathbf{>50}$ K}}&\textbf{Overall}\\
& Prec. & Recall & F1 &\textbf{F1}\\
\hline
{Ungrouped} & 0.80 & 0.48 & 0.60&0.60\\
\hline
{Grouped} & 0.98 & 0.62 & 0.76 &0.76\\
\hline
\end{tabular}
\caption{Comparison of XGBoost classification performance for minority class using grouped and ungrouped CSV-style prompt design for the Adult Income dataset.}
\label{tab:AdultIncomePromptDesign}
\end{table}

\subsection{Determining Correlation Threshold}
We first determine the correlation between each imputation feature (feature with missing values) and other columns in the dataset. We use Pearson correlation between two numerical features, Cramer's V between categorical variables and eta ($\eta$) correlation ratio between a categorical and numerical variable. We combine the absolute values of all correlations into a single list, rank them in descending order and plot them to determine the ``elbow point" in the plot. This is the point where the correlation values drop sharply, and then the rate of decrease becomes notably less steep. The correlation value at this elbow point is a lower bound for our threshold. After repeating this process for all imputation features, we can either choose the minimum correlation threshold or just keep different thresholds for different features. If the correlation values are approximately equal, we assign this approximate value to the threshold. The way we combine the correlation thresholds of different features depends on the dataset under consideration. After obtaining the overall threshold, we select all columns having higher correlation with the imputation feature than the threshold value. We repeat the experiment with another value below the obtained threshold for comparison. 

\subsection{Data Imputation Technique} 
We impute all features with missing entries considering one feature at a time. The technique is designed to generate high-quality data to impute missing values in such a way that the input prompt length is optimized. One known fact is that reducing the prompt length decreases the computational costs of the LLM and enables more information to be fed in for the same token size constraints. In this paper, we show that by discarding dataset features that exhibit weak associations with the feature being imputed and focusing on the most relevant information, we can achieve an imputation quality atleast as good as using the complete dataset information.

Once we fix the correlation threshold and select the important columns for each imputation feature, we determine the imputation feature that is considered relevant for the highest number of other imputation features. This approach allows us to add increasingly informative context to the prompt as we impute more features, thereby enhancing the imputation process for the remaining features.

Once the order of feature imputation is determined, we instruct the LLM to impute missing values in the first feature column using a group-wise CSV-style prompt. The LLM outputs the completed feature column after imputing missing values. Then, we update the prompt by replacing the original feature column with missing entries by the completed column outputted by the LLM. Then, we use this updated prompt for imputing the next feature. We continue to update the prompt after imputing each feature to add more context for the LLM to perform the next imputation. The process continues until we impute all the missing values in the dataset.

\begin{table}[ht]
\centering
\setlength{\tabcolsep}{2pt}
\small
\begin{tabular}{|l|l|ccc|ccc|}
\hline
\textbf{Model used} & \textbf{Correlation} & \multicolumn{3}{c|}{\textbf{Target-0}} & \multicolumn{3}{c|}{\textbf{Target-1}} \\
 & \textbf{Threshold} & Prec. & Recall & F1 & Prec. & Recall & F1 \\
\hline
\multirow{3}{*}{XGBoost} & 0 & 0.67 & 0.97 & 0.79 & 0.94 & 0.52 & 0.67 \\
 & 0.15 & 0.69 & 0.98 & 0.81 & 0.97 & 0.57 & 0.72 \\
 & 0.2 & 0.70 & 0.98 & 0.82 & 0.97 & 0.58 & 0.73 \\
\hline
\multirow{3}{*}{Random Forest} & 0 & 0.68 & 0.90 & 0.77 & 0.85 & 0.57 & 0.68 \\
& 0.15 & 0.70 & 0.97 & 0.81 & 0.95 & 0.58 & 0.72 \\
 & 0.2 & 0.70 & 0.97 & 0.81 & 0.95 & 0.58 & 0.72 \\
\hline
\end{tabular}
\caption{Classification performance by target class and model for the Travel dataset}
\label{tab:TravelClasswise}
\end{table}
\section{Experiments and Insights}
We use the latest \textbf{GPT 4.1} model to impute missing values in both datasets. An example prompt to perform this task is shown in Table~\ref{tab:PromptExample}. Across all the experiments, we report results from two ML classifiers: \emph{XGBoost} and \emph{Random forest classifier}. These classifiers are trained on the original datasets (with missing rows dropped) and tested on the set of imputed records. We compare the quality of imputation for three correlation threshold values for each dataset as assessed by ML classification performance. The imputation method is evaluated using F1 scores, precision, recall, balanced accuracy (BAL ACC) and ROC AUC scores. 
We first report the results from our experiments evaluating how prompt design influences the quality of data imputation as evidenced through classification-based evaluation metrics. Then, we investigate how feature space reduction influences imputation quality.
\subsection{Exploring the importance of prompt design}
We investigate the importance of the CSV-style group-wise prompting method for imputing missing values by comparing it to a CSV-style prompt with the same number of completed examples and missing records, but with no clear class separation. The example records for the ungrouped-example prompt are obtained by random sampling from the completed records of the original dataset. The precision, recall and F1 scores for the minority class and the overall F1 scores obtained using XGBoost classifier are reported in Table~\ref{tab:AdultIncomePromptDesign}. For our experiment, we have trained the model on the original Adult Income dataset and tested it on the set of imputed records. We observe a notable increase in the the F1 scores for the minority class and a boost in the overall performance of the model for group-wise CSV style prompt as compared to the ungrouped CSV-style prompt. 


\begin{table}[th]
\centering
\setlength{\tabcolsep}{4pt}
\small
\begin{tabular}{|l|l|c|c|c|}
\hline
\multirow{2}{*}\textbf{Model Used}& \textbf{Corr.} & \textbf{BAL ACC} & \textbf{F1} & \textbf{ROC AUC}\\
& \textbf{Thres.}&&&\\
\hline
\multirow{3}{*}{XGBoost} & 0 & 0.742 & 0.667 & 0.92\\
& 0.15 & 0.775 & 0.716  & 0.95\\
& 0.2 & 0.783 & 0.729 & 0.96\\
\hline
\multirow{3}{*}{Random Forest} & 0 & 0.733 & 0.680 & 0.91\\
& 0.15 & 0.775 & 0.722 & 0.95\\
& 0.2 & 0.775 & 0.722 & 0.96\\
\hline
\end{tabular}
\caption{Overall XGBoost and Random Forest Classifier performance metrics for Travel dataset; Columns in the table represent the Model Used, Correlation Threshold, Balanced accuracy, F1 score and ROC AUC.}
\label{Travel: Overall metrics}
\end{table}

\begin{figure}[tb]
    \centering
    \includegraphics[width=0.77\linewidth]{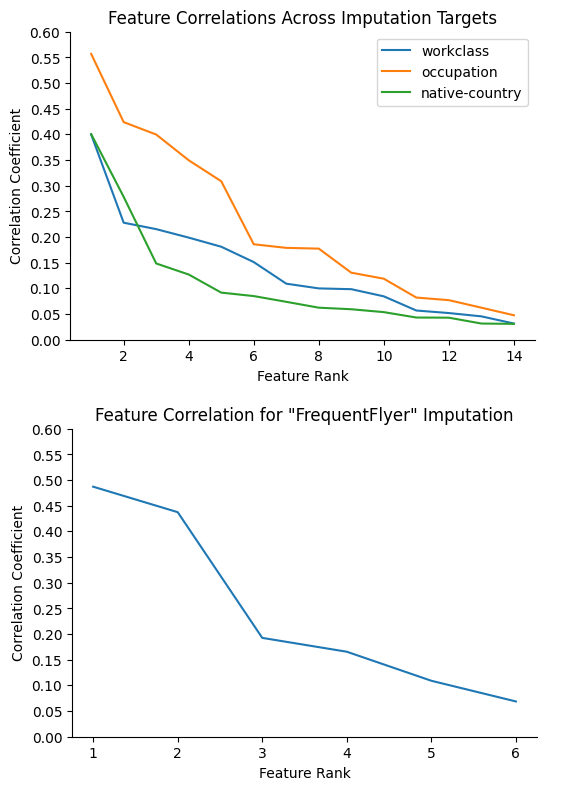}
        \caption{Elbow Plot illustrating the selection of a correlation threshold for imputing missing entries in the Adult Income and Travel datasets. The plot shows sorted absolute correlation values of the imputations features with the potential predictor features in the dataset.}
    \label{fig:ElbowAnalysisTravelandAdultIncome}
\end{figure}

\textbf{Estimating the correlation threshold on our example datasets}: We evaluate the correlation between each of the imputation features with all other features in the dataset using Cramer's V for categorical features and eta correlation ratio for numerical features. The plots of absolute correlation values are shown in Figure~\ref{fig:ElbowAnalysisTravelandAdultIncome}. For the Adult Income dataset, the imputation features are `workclass', `occupation' and `native-country'. For this dataset, we observe that the elbow point is between $0.20$ and $0.25$ for `workclass' feature, around $0.20$ for `occupation' and around $0.15$ for `native-country'. We observe that for `native-country', applying a correlation cutoff of $0.2$ for feature selection produces the same set of features as applying a cutoff of $0.15$. Thus, we consider a correlation threshold of $0.2$ in our experiments. For the Travel dataset, the imputation feature is `FrequentFlyer'. From the elbow plot showing the correlation of `FrequentFlyer' with other features, we observe that the elbow point lies between $0.15$ and $0.20$. Thus, we select $0.2$ as the optimal correlation threshold for this dataset too. For both datasets, we also obtain the imputed datasets by considering all features in the prompt and with another correlation cutoff below the obtained threshold to understand the effect of including noisy features on the ML classification performance. 

\begin{table}[tp]
\centering
\setlength{\tabcolsep}{2pt}
\small
\begin{tabular}{|l|l|ccc|ccc|}
\hline
\textbf{Model used} & \textbf{Correlation} & \multicolumn{3}{c|}{\textbf{Income $\mathbf{<=50}$ K}} & \multicolumn{3}{c|}{\textbf{Income $\mathbf{>50}$ K}} \\
 & \textbf{Threshold} & Prec. & Recall & F1 & Prec. & Recall & F1 \\
\hline
\multirow{3}{*}{XGBoost} & 0 & 0.73 & 0.99 & 0.84 & 0.99 & 0.64 & 0.77 \\
 & 0.1 & 0.72 & 0.99 & 0.83 & 0.98 & 0.61 & 0.75 \\
 & 0.2 & 0.71 & 0.99 & 0.83 & 0.98 & 0.60 & 0.74 \\
\hline
\multirow{3}{*}{Random Forest} & 0 & 0.74 & 0.98 & 0.84 & 0.97 & 0.65 & 0.78 \\
& 0.1 & 0.71 & 0.98 & 0.83 & 0.97 & 0.61 & 0.75 \\
 & 0.2 & 0.71 & 0.98 & 0.82 & 0.97 & 0.60 & 0.74 \\
\hline
\end{tabular}
\caption{Classification performance by target class and model for the Adult Income dataset}
\label{tab:AdultIncomeClasswise}
\end{table}

\subsection{Exploring the effect of token reduction on ML classification performance}
We evaluate the quality of imputation by assessing the classification performance when ensemble learning models are trained on the original datasets and tested on the set of imputed records. The models used for our investigations are -- \emph{XGBoost} and \emph{Random Forest} Classifiers. We perform the experiments for three correlation thresholds for each dataset. The number of columns included in the prompt for imputing missing values in features corresponding to each correlation threshold is presented in Table~\ref{tab:AdultIncomeFeatureReduction} and Table~\ref{tab:TravelFeatureReduction}. As shown in Table~\ref{tab:AdultIncomeClasswise}, the F1 scores for the minority class are maintained (with a maximum difference of $0.04$) despite a significant reduction in number of input features used in the prompt (see Table~\ref{tab:AdultIncomeFeatureReduction}). Also, the balanced accuracy scores just drop by about $2.7\%$ despite aggressively shrinking feature information by about $76.19\%$ for both XGBoost and Random Forest Classifiers. Also, the models maintain nearly equal ROC AUC scores for all thresholds.

For Travel, omitting irrelevant features from the prompt actually boosts the imputation quality for the minority class as seen from the jump in F1 scores by about $7.46\%$ (Table \ref{tab:TravelClasswise}), an increase in overall ROC AUC scores from around $0.91$ to $0.95$ (Table \ref{Travel: Overall metrics}) and a rise in balanced accuracy scores by $4.45\%$ (Table \ref{Travel: Overall metrics}) when the correlation threshold is increased to $0.15$. Thus, the group-wise and overall metrics show significant improvement as we eliminate unnecessary information from the prompt for the imputation task. These results suggest that LLM-driven prompt modifications significantly influence classification performance in small data setting yielding an improved set of scores. However, in larger datasets this effect is less visible. This may be due to the model's reliance on a substantial sample of data that enables it to filter out irrelevant information by itself, consequently leading to lesser effect on performance scores. 

\begin{table}[ph]
\centering
\setlength{\tabcolsep}{4pt}
\small
\begin{tabular}{|l|l|c|c|c|}
\hline
\multirow{2}{*}\textbf{Model Used}& \textbf{Corr.} & \textbf{BAL ACC} & \textbf{F1} & \textbf{ROC AUC}\\
&\textbf{Thres.}&&&\\
\hline
\multirow{3}{*}{XGBoost} & 0 & 0.814 & 0.774 & 0.96\\
& 0.1 & 0.799 & 0.752  & 0.96\\
& 0.2 & 0.792 & 0.743 & 0.96\\
\hline
\multirow{3}{*}{Random Forest} & 0 & 0.813 & 0.776 & 0.95\\
& 0.1 & 0.794 & 0.748 & 0.94\\
& 0.2 & 0.790 & 0.740 & 0.94\\
\hline
\end{tabular}
\caption{Overall XGBoost and Random Forest Classifier performance metrics for Adult Income dataset}
\label{Adult Income:Overall Metrics}
\end{table}

\begin{table}[tp]
\centering
\setlength{\tabcolsep}{3pt}
\small
\begin{tabular}{|l|ccc|l|}
\hline
\textbf{Corr.} & \multicolumn{3}{c|}{\textbf{No. of Col Retained for Imputation}} & \textbf{\% Reduction in}\\
\textbf{Thres.} & Workclass & Occupation & Native-country &\textbf{Feature Space}\\
\hline
0 & 14/14 & 14/14 & 14/14&-\\
0.1 & 7/14 & 10/14 & 4/14&50\%\\
0.2 & 3/14 & 5/14 & 2/14&76.19\%\\
\hline
\end{tabular}
\caption{Column retention counts (No. of columns retained / Total no. of columns) for imputation of 'workclass','occupation', and 'native-country' features in the Adult Income dataset. Percentage reduction in feature space is the ratio between the total number of columns removed at each non-zero correlation threshold and the total number of features at zero threshold, illustrating how increasing the correlation cutoff progressively filters out less correlated features and reduces dimensionality.}
\label{tab:AdultIncomeFeatureReduction}
\end{table}

\begin{table}[tp]
\centering
\setlength{\tabcolsep}{2pt}
\small
\begin{tabular}{|l|c|c|}
\hline
\textbf{Corr.} & \textbf{No. of Col. Retained for Imputation} &\textbf{\% Reduction in}\\
\textbf{Thres.}&&\textbf{Feature Space}\\
\hline
0 & 6/6 & -\\
0.15 & 4/6 & 33.33\%\\
0.2 & 2/6 & 66.67\%\\
\hline
\end{tabular}
\caption{Column retention counts (No. of columns retained / Total no. of columns) for imputation of `FrequentFlyer' feature in the Travel dataset. Percentage reduction in feature space is the ratio between the total number of columns removed at each non-zero correlation threshold and the total number of features at zero threshold.}
\label{tab:TravelFeatureReduction}
\end{table}

\section{Discussion}
There are several advantages of leveraging CSV-style prompting. Firstly, for \emph{minimum data preprocessing}: This prompting approach eliminates the need for data preprocessing, retaining the original column names and formats. This strategy also preserves the integrity of raw data, enabling the inclusion of both categorical and numerical variables without the need for extensive modifications~\cite{kim2024epic}; Secondly, \emph{optimized token usage}: Our approach optimizes token consumption for each data value by representing tabular data within prompts using a CSV-style format. This format is especially beneficial since it permits a greater number of in-context learning instances within the same token limits, which is especially helpful given the constrained context window of LLMs~\cite{kim2024epic}; Thirdly, another advantage of this method is that the LLM encounters both the majority and minority class samples in a proportionate manner so it understands the trends within each group effectively before imputing samples from each group. This is in contrast to many of the fine-tuning methods, which tend to overfit the majority class values. Finally, the lower the prompt's token size, more data samples can be included as examples for the LLM to analyze. This is important as we want the prompt samples to represent the original data distribution. The LLMs ability to impute missing values is confined to its analysis of examples in the input.\\ 

\noindent
Through our experimental evaluation, we also found that LLM-based prompt modifications significantly influence the classification performance in case of small-sized datasets yielding an improved set of scores when irrelevant features were ignored in the prompt. However, in the case of larger datasets, this effect is relatively less visible. This may be due to the fact that when there is a larger sample of training data, the model may filter out irrelevant information by itself. Thus, there is a relatively lower impact on the performance scores. These observations lead to our next set of questions 1) In what scenarios does prompting amplify or suppress biases present in the original data via its own training? 2) if the prompt has a larger influence on datasets of smaller sizes, can we guarantee that the synthetic data generation process could be reproducible? There will be major ethical issues that may lead to practical consequences in domains such as health, legal, or finance. Thus, it is crucial to further explore the impact of prompting on synthetic data generation and imputation in sensitive domains. 

\noindent
\textbf{Limitations} We tested our technique on two datasets with imputation features being categorical. Thus, the results from our experiments may not be immediately generalized to imputation features of all types. Another potential issue could be handling datasets with a majority of the imputation feature correlations having similar values as obtaining a correlation threshold that can effectively filter out noisy component features can be challenging. Future research will include testing this method with datasets from various domains containing both numerical and categorical imputation features and exploring ways to determine an optimal correlation threshold in difficult scenarios.  

\section{Conclusion}
The approach discussed in this paper leverages prompting techniques and the power of feature correlation to develop a tabular data imputation method optimized for input and output token usage especially for imbalanced datasets. With the help of a group-wise CSV-based prompting method and the dataset's inter-feature associations, this work is a step towards leveraging the in-context learning capabilities of LLMs for synthetic data generation for class-imbalance problems. In this work, we tested our method on two binary classification datasets and evaluated the imputation quality using classification performance by building ensemble learning models. From our observations, we conclude that this method gives the user an opportunity to strike a balance between a desired level of accuracy and token consumption depending on the use-case. In our tests, we also observed that removing irrelevant information from the prompt can boost imputation quality especially for small-sized datasets.

\bibliography{references}

\newpage 
\section{Appendix}
\begin{table*}
\begin{tabular}{|l|l|l|}
\hline \multicolumn{2}{|c|}{Template} & Prompt sample \\
\hline \multicolumn{2}{|r|}{\multirow{6}{*}{Descriptions}} & \begin{tabular}{l}
Churn: whether customer churns or doesnt churn for tour and travels company, \\
age: the age of customer, \\
FrequentFlyer: whether customer takes frequent flights, \\
AnnualIncomeClass: class of annual income of user, \\
ServicesOpted: number of times services opted during recent years, \\
AccountSyncedToSocialMedia: whether company account of user synchronised to their social media,
\end{tabular} \\
\hline Set & Header & \begin{tabular}{l}
Churn, Age, FrequentFlyer, AnnualIncomeClass, ServicesOpted, \\
AccountSyncedToSocialMedia, BookedHotelOrNot
\end{tabular} \\
\hline & Group & \begin{tabular}{l}
A. \\
Churn, 28, Yes, High Income, 6, No, Yes \\
Churn, 37, Yes, Low Income, 4, Yes, Yes \\
Churn, 30, Yes,Low Income, 1,Yes,Yes
\end{tabular} \\
\hline & Group & \begin{tabular}{l}
B. \\
Doesnt churn, 38, No, Low Income,1,Yes,No \\
Doesnt churn, 28, No Record,Low Income,5,No,Yes \\
Doesnt churn, 34, Yes, Low Income, 1,No,No
\end{tabular} \\
\hline Set & Header & \begin{tabular}{l}
Churn, Age, FrequentFlyer, AnnualIncomeClass, ServicesOpted, \\
AccountSyncedToSocialMedia, BookedHotelOrNot
\end{tabular} \\
\hline & Group & \begin{tabular}{l}
A. \\
Churn, 29, Yes, Low Income, 6, No, No \\
Churn, 37, Yes, High Income, 4, Yes, Yes \\
Churn, 25, Yes,Low Income, 2,Yes,Yes
\end{tabular} \\
\hline & Group & \begin{tabular}{l}
B. \\
Doesnt churn, 33, Yes, High Income,1,Yes,No \\
Doesnt churn, 30,Yes,Low Income,5,No,Yes \\
Doesnt churn, 34,No, Low Income, 1,No,Yes
\end{tabular}\\
\hline & & \begin{tabular}{l}
Given the above data, fill in the missing values in the data sample below:
\end{tabular} \\
\hline & Group & \begin{tabular}{l}
A. \\
Churn, 28, Yes, No record, 6, No, Yes \\
Churn, 37, Yes, Low Income, 4,No Record, Yes \\
Churn, 30,No Record,Low Income, 1,Yes,Yes
\end{tabular} \\
\hline & Group & \begin{tabular}{l}
B. \\
Doesnt churn, 38, No, Low Income,1,Yes,No Record \\
Doesnt churn, 28, No Record,Low Income,5,No Record,Yes \\
Doesnt churn, 34,No Record, Low Income, 1,No,No
\end{tabular}\\
\hline
\end{tabular}
\caption{Example of a group-wise CSV-style prompt for the Travel dataset. This prompt contains two sets of completed samples to impute one set of missing records with the same sample size. The completed records are extracted by random sampling from both groups.}
\label{tab:PromptExample}
\end{table*}

\end{document}